\documentclass[5p]{elsarticle}

\usepackage{lineno,hyperref}
\modulolinenumbers[5]

\usepackage{amsmath} 
\usepackage{amssymb} 
\usepackage{xspace} 
\usepackage[caption=false]{subfig} 

\usepackage{tabularx} 
\usepackage{booktabs} 
\usepackage{multirow} 

\usepackage{algorithm}
\usepackage[noend]{algpseudocode}

\journal{Computer Vision and Image Understanding}

\def\etal{\emph{et al.}\@\xspace}
\def\ie{\emph{i.e.}\@\xspace}
\def\cf{\emph{cf.}\@\xspace}
\newcommand{\argmax}{\operatornamewithlimits{arg\,max}}
\newcommand{\argmin}{\operatornamewithlimits{arg\,min}}
\newcommand{\softmax}{\mathrm{softmax}}
\renewcommand{\vec}[1]{\mathbf{#1}}
\newcommand{\mat}[1]{\mathbf{#1}}

\def\IR{\mathbb{R}}

\begin{document}

\begin{frontmatter}

\title{A Bag-of-Words Equivalent Recurrent Neural Network for Action Recognition}

\author{Alexander Richard, Juergen Gall}
\address{University of Bonn, R\"omerstra{\ss}e 164, 53177 Bonn, Germany \\
         \{richard,gall\}@iai.uni-bonn.de}


\begin{abstract}
The traditional bag-of-words approach has found a wide range of applications in
computer vision. The standard pipeline consists of a generation of a visual
vocabulary, a quantization of the
features into histograms of visual words, and a classification step for which
usually a support vector machine in combination with a non-linear kernel is used.
Given large amounts of data, however, the model suffers from a lack of discriminative
power. This applies particularly for action recognition, where the vast amount of
video features needs to be subsampled for unsupervised visual vocabulary generation.
Moreover, the kernel computation can be very expensive on large datasets.
In this work, we propose a recurrent neural network that is equivalent to the
traditional bag-of-words approach but enables for the application of discriminative
training. The model further allows to incorporate the kernel computation into the
neural network directly, solving the complexity issue and allowing to represent the
complete classification system within a single network.
We evaluate our method on four recent action recognition benchmarks and show that
the conventional model as well as sparse coding methods are outperformed.
\end{abstract}




\end{frontmatter}


\section{Introduction}

The traditional bag-of-words or bag-of-features model has been of interest for
numerous computer vision tasks ranging from object classification~\cite{csurka2004visual}
and discovery~\cite{sivic2005discovering} to texture classification~\cite{zhang2007local}
and object retrieval~\cite{VideoGoogle}. Although recently other approaches like
convolutional neural networks~\cite{krizhevsky2012imagenet}
or Fisher vectors~\cite{perronnin2010improving} show better classification results,
bag-of-words models still experience great popularity due to their simplicity and computational
efficiency. Particularly for the task of action recognition, where state-of-the-art
feature extraction approaches lead to a vast amount of features even for comparably
small datasets~\cite{wang2013dense, wang2013action}, compact and efficient feature
representations like bag-of-words are widely used
\cite{taralova2014motion,wang2013dense,reddy2013recognizing,jhuang2013jhmdb,peng2014bag}.

For most classification tasks, the application of a bag-of-words model can be
subdivided into three major steps. First, a visual vocabulary is created by
clustering the features, usually using kMeans or a Gaussian mixture model.
In a second step, the input data is quantized and eventually represented
by a histogram of the previously obtained visual words. Finally, the data is
classified using a support vector machine.

However, there are some significant drawbacks in this pipeline. Clustering algorithms
like kMeans and Gaussian mixture models require the computation of distances of
all input features to all cluster centers in each iteration. Due to the extensive amount
of features generated by state-of-the-art feature extraction algorithms for videos such
as improved dense trajectories~\cite{wang2013action}, it is infeasible to
run the algorithms on the complete data and a subset has to be sparsely sampled.
Although the authors of \cite{wang2013dense} propose to run kMeans several times
and select the most compact clustering to ensure a good subsampling, kMeans is
usually not sensitive to the subsampling of local descriptors.
However, the visual vocabulary is created without supervision
and optimized to be a good representation of the overall data, whereas for
classification, the visual vocabulary should ideally be optimized to best separate
the classes. In the actual classification step, a non-linear kernel can be
applied to the data to increase the accuracy. While this is not critical
for small datasets, it can become infeasible for large scale tasks since the kernel
computation is quadratic in the amount of training instances.

In this work, we present a novel approach to model a visual vocabulary and the actual
classifier directly within a recurrent neural network that can be shown to be
equivalent to the bag-of-words model. The work is based on the preliminary work
\cite{richard2015bow}, where we already showed how to build a bag-of-words equivalent
recurrent neural network and learn a visual vocabulary by optimizing the class
posterior probabilities directly rather than optimizing the sum of squared distances
as in kMeans. This way, we compensate for the lack of discriminative power in traditionally
learned vocabularies. We extend our proposed
model from~\cite{richard2015bow} such that the kernel computation can be
included as a neural network layer. This resolves the complexity issues for large
scale tasks and allows to use the neural network posteriors for classification instead
of an externally trained support vector machine.

For evaluation, we use the recognition framework proposed in
\cite{wang2013dense,wang2013action} and replace the kMeans-based bag-of-words model by
our recurrent neural network.
We analyze our method on four recent action recognition benchmarks and show that
our model constantly outperforms the traditional bag-of-words approach. Particularly
on large datasets, we show an improvement by two to five percentage points while it
is possible to reduce the number of extracted features considerably.
We further compare our method to state-of-the-art feature encodings that are
widely used in action recognition and image classification.


\section{Related work}

In recent years, many approaches have been developed to improve the traditional
bag-of-words model.
In \cite{perronnin2006adapted}, class-dependent and universal visual vocabularies
are computed. Histograms based on both vocabularies are then merged and used to
train a support vector machine for each class. Perronnin \etal argue that additional
discriminativity is added to the model via the class-dependent vocabularies.
In \cite{cai2010learning}, a weighted codebook is generated using metric learning.
The weights are learned such that the similarity between instances of the same class
is larger than between instances of different classes.
Lian \etal \cite{lian2010probabilistic} combine an unsupervised generated vocabulary
with a supervised logistic regression model and outperform traditional models.
Sparse coding has successfully been applied in \cite{yang2009linear} and
\cite{wang2010locality}. The first extends spatial pyramid matching by a sparse coding step,
while the latter, termed locality constrained linear coding, projects a descriptor
into a space defined by its $ K $ nearest codewords.

Goh \etal \cite{goh2012unsupervised} use a restricted
Boltzman machine in order to learn a sparsely coded dictionary. Moreover, they show that
the obtained dictionary can be further refined with supervised fine-tuning. To this end,
each descriptor is assigned the label of the corresponding image. Note that our approach,
in contrast, allows for supervised training beyond descriptor level and on video level
directly. Similar to our method, the authors of \cite{boureau2010learning} also apply
supervised dictionary learning on video level and show that it can improve sparse coding
methods. However, while their approach is limited to optimizing a dictionary in combination
with a linear classifier and logistic loss, our method allows for the direct
application of neural network optimization methods, and allows to include feature
mappings that correspond to the application of various non-linear kernels.
Yet, neither of the approaches has been applied to action recognition.

Recently, in action recognition, more sophisticated feature encodings such as
VLAD \cite{jegou2012aggregating} or improved Fisher vectors \cite{perronnin2010improving}
gained attention. Particularly Fisher vectors are used in most action recognition
systems \cite{wang2013action,peng2014action, oneata2013action}. They usually
outperform standard bag-of-words models but have higher computation and memory
requirements.
In \cite{simonyan2013deep}, Fisher vectors are stacked in multiple layers and for
each layer, discriminative training is applied to learn a dimension reduction,
whereas the authors of \cite{wang2016mofap} develop a hierarchical system of motion
atoms and motion phrases based on Fisher vectors amongst other motion features.
Ni \etal~\cite{ni2015motion} showed that clustering dense trajectories
into motion parts and generating discriminative weighted Fisher vectors can
further improve action recognition.
In order to boost the performance of VLAD, Peng \etal apply supervised dictionary
learning \cite{peng2014boosting}. They propose an iterative optimization scheme that
alters between optimizing the dictionary while keeping the classifier weights fixed
and optimizing the classifier weights while keeping the dictionary fixed, respectively.
Investigating structural similarities of neural networks and Fisher vectors,
Sydorov \etal \cite{sydorov2014deep} propose a similarly alternating optimization
scheme in order to train a classifier and the parameters of a Gaussian mixture
model used for the Fisher vectors.
While both of these methods greedily optimize the non-fixed part of the respective
model in each step, our approach avoids such an alternating greedy scheme and
allows for unconstrained optimization of all model parameters at once.

Due to the remarkable success of convolutional neural networks
(CNNs) for image classification \cite{krizhevsky2012imagenet}, neural networks
recently also experience a great popularity in action recognition.
On a set of over one million Youtube videos with weakly annotated class labels,
Karpathy \etal \cite{karpathy2014large} successfully trained a CNN and showed
that the obtained features also perform reasonably well on other action recognition
benchmarks. Considering the difficulty of modeling movements in a simple CNN,
the authors of \cite{simonyan2014two} introduced a two-stream CNN processing not
only single video frames but also the optical flow as additional input. Their
results outperform preceding CNN based approaches, underlining the importance
of motion features for action recognition.
In \cite{donahue2015long} and \cite{srivastava2015unsupervised}, CNN features are
used in LSTM networks in order to explore temporal information. These approaches,
however, are not yet competitive to state-of-the-art action recognition systems
such as \cite{simonyan2014two} or \cite{wang2013action}.
Although improved dense trajectories \cite{wang2013action} are still the de-facto
state-of-the-art for action recognition, Jain \etal recently showed that they
can be complemented by CNN features \cite{jain2015objects}.
In another approach, the authors of~\cite{wang2015tdd} propose a combination of dense
trajectories with the two-stream CNN~\cite{simonyan2014two} in order to obtain
trajectory pooled descriptors.

Our model aims at closing the gap between the two coexisting approaches of traditional
models like bag-of-words on the one hand and deep learning on the other hand.
Designing a neural network that can be proven equivalent to the traditional
bag-of-words pipeline including a non-linear kernel and support vector machine, we
provide insight into relations between both approaches. Moreover, our framework
can easily be extended to other tasks by exchanging or adding specific layers to the
proposed neural network. For better reproducibility, we made our source code available.
\footnote{\texttt{https://github.com/alexanderrichard/squirrel}}




\section{Bag-of-Words Model as Neural Network}
\label{sec:technical}

In this section, we first define the standard bag-of-words model and propose a
neural network representation. We then discuss the equivalence of both models.

\subsection{Bag-of-Words Model}

Let $ \vec{x} = (x_1,\dots,x_T) $ be a sequence of $ D $-dimensional feature
vectors $ x_i \in \IR^D $ extracted from some video and
$ \mathcal{C} = \lbrace 1,\dots,C \rbrace $ the set of classes.
Further, assume the training data is given as $ \lbrace (\vec{x}_1,c_1),\dots,(\vec{x}_N,c_N) \rbrace $.
In the case of action recognition, for example, each observation $ \vec{x}_i $
is a sequence of feature vectors extracted from a video and $ c_i $ is the action
class of the video.
Note that the sequences usually have different lengths, \ie for two different
observations $ \vec{x}_i $ and $ \vec{x}_j $, usually $ T_i \neq T_j $.

The objective of a bag-of-words model is to quantize each observation
$ \vec{x} $ using a fixed vocabulary of $ M $ visual words,
$ \mathcal{V} = \lbrace v_1,\dots,v_M \rbrace \subset \IR^D $.
To this end, each sequence is represented as a histogram of posterior probabilities $ p(v|x) $,
\begin{align}
    \mathcal{H}(\vec{x}) = \frac{1}{T} \sum_{t=1}^{T} h(x_t),
    \quad
     h(x_t) = \left( \begin{array}{c}
                        p(v_1|x_t) \\
                        \vdots \\
                        p(v_M|x_t)
                     \end{array} \right).
    \label{histogram}
\end{align}
Frequently, kMeans is used to generate the visual vocabulary. In this case,
$ h(x_t) $ is a unit vector, \ie the closest visual word has probability one and
all other visual words have probability zero. Based on the histograms, a
probability distribution $ p(c|\mathcal{H}(\vec{x})) $ can be modeled. Typically, a
support vector machine in combination with a non-linear kernel is used for
classification.

\subsection{Conversion into a Neural Network}
\label{sec:conversionIntoNN}

The result of kMeans can be seen as a mixture distribution describing the
structure of the input space. Such distributions can also be modeled with
neural networks. In the following, we propose a transformation of the bag-of-words
model into a neural network.

The nearest visual word $ \hat v = \argmin_m \|x-v_m\|^2 $ for a feature vector $ x $ can be seen as the maximizing argument of the posterior form of a Gaussian distribution,
\begin{align}
    p_{\mathrm{KM}}(v_m|x) &= \frac{p(v_m)p(x|v_m)}{\sum_{\tilde m}p(v_{\tilde m})p(x|v_{\tilde m})} \\
                           &= \frac{\exp{\big(-\frac{1}{2}(x-v_m)^\intercal(x-v_m)\big)}}
                                  {\sum_{\tilde m}\exp{\big(-\frac{1}{2}(x-v_{\tilde m})^\intercal(x-v_{\tilde m})\big)}},
    \label{visualWordsPosterior}
\end{align}
assuming a uniform prior $ p(v_m) $ and a normal distribution
$ p(x|v_m) = \mathcal{N}(x|v_m, \mat{I}) $ with mean $ v_m $ and unit variance.
Using maximum approximation, \ie shifting all probability mass to the most
likely visual word, a probabilistic interpretation for kMeans can be obtained:
\begin{align}
    \hat p_{\mathrm{KM}}(v_m|x) =
        \left \lbrace \begin{array}{rl}
            1, & \text{ if } v_m = \argmax_{\tilde m} p_{\mathrm{KM}}(v_{\tilde m}|x), \\
            0, & \text{ otherwise.}
        \end{array} \right. \label{probKMeans}
\end{align}
Inserting  $ \hat p_{\mathrm{KM}}(v_m|x) $ into the histogram equation \eqref{histogram} 
is equivalent to counting how often each visual word $ v_m $ is the nearest representative for the
feature vectors $ x_1,\dots,x_T $ of a sequence $ \mathbf{x} $.

Now, consider a single-layer neural network with input $ x \in \IR^D $ and $ M $-dimensional $ \softmax $
output that defines the posterior distribution
\begin{align}
    p_{\mathrm{NN}}(v_m|x) :=& \ \softmax_m(\mat{W}^\intercal x + b) \\
                            =& \ \frac{\exp{ \big( \sum_d w_{d,m} x_d + b_m \big) }}{\sum_{\tilde m} \exp{ \big( \sum_d w_{d,\tilde m} x_d + b_{\tilde m} \big) }}
    , \label{nnPosteriors}
\end{align}
where $ \mat{W} \in \IR^{D \times M} $ is a weight matrix and $ b \in \IR^M $
the bias. With the definition
\begin{align}
    \mat{W} &= (v_1 \dots v_M), \label{nnWeights} \\
    b &= -\frac{1}{2}(v_1^\intercal v_1 \dots v_M^\intercal v_M)^\intercal, \label{nnBias}
\end{align}
an expansion of Equation \eqref{visualWordsPosterior} reveals that
\begin{align}
    p_{\mathrm{NN}}(v_m|x) = \frac{\exp{ \big( -\frac{1}{2}v_m^\intercal v_m + v_m^\intercal x \big) }}{\sum_{\tilde m} \exp{ \big( -\frac{1}{2}v_{\tilde m}^\intercal v_{\tilde m} + v_{\tilde m}^\intercal x} \big) }
                           = p_{\mathrm{KM}}(v_m|x).
\end{align}

A recurrent layer without bias and with unit matrix as weights for both the
incoming and recurrent connection is added to realize the summation over
the posteriors $ p_{\mathrm{NN}}(v_m|x) $ for the histogram computation, \cf
Equation \eqref{histogram}. The histogram normalization is achieved using
the activation function
\begin{align}
    \sigma_t(z) = \left\lbrace \begin{array}{rl}
                                   z & \text{ if } t < T, \\
                                   \frac{1}{T} z & \text{ if } t = T.
                               \end{array} \right.
\end{align}
Given an input sequence $ \vec{x} $ of length $ T $, the output of the recurrent
layer is
\begin{align}
      & \sigma_T\Big(h(x_T) + \sigma_{T-1}\big(h(x_{T-1}) +\sigma_{T-2}(h(x_{T-2}) + \dots)\big)\Big) \nonumber \\
    = & \frac{1}{T} \sum_{t=1}^T h(x_t) = \mathcal{H}(\vec{x}).
\end{align}

\begin{figure}[t]
\begin{center}
    \includegraphics{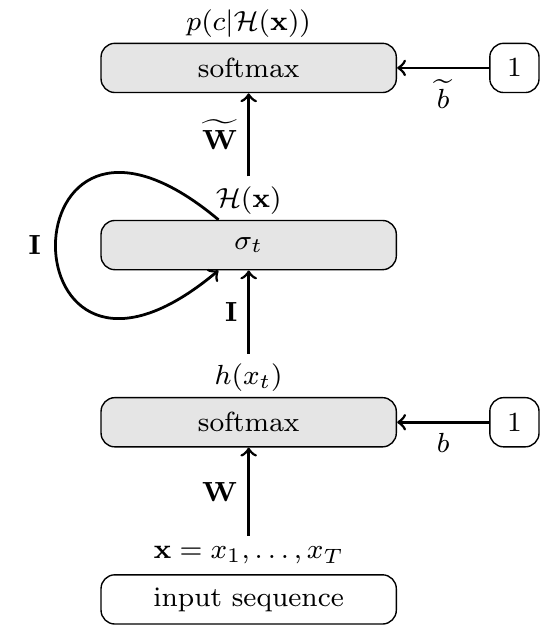}
\end{center}
\caption{Neural network encoding the bag-of-words model. The output layer is
         discarded after training and the histograms from the recurrent layer
         are used for classification in combination with a support vector machine.}
\label{fig:nn}
\end{figure}

So far, the neural network computes the histograms $ \mathcal{H}(\vec{x}) $ for
given visual words $ v_1,\dots,v_M $. In order to train the visual words
discriminatively and from scratch, an additional $ \softmax $ layer with $ C $
output units is added to model the class posterior distribution
\begin{align}
    p(c|\mathcal{H}(\vec{x})) = \softmax_c(\mat{\widetilde W}^\intercal \mathcal{H}(\vec{x}) + \widetilde b).
\end{align}
It acts as a linear classifier on the histograms and allows for the application
of standard neural network optimization methods for the joint estimation of the visual
words and classifier weights. Once the network is trained, the $ \softmax $
output layer can be discarded and the output of the recurrent layer is used as
histogram representation. The complete neural network is depicted in Figure
\ref{fig:nn}. 

Note the difference of our method to other supervised learning methods like the
restricted Boltzman machine of~\cite{goh2012unsupervised}. Usually,
each feature vector $ x_t $ extracted from a video gets assigned the class of the
respective video. Then, the codebook is optimized to distinguish the classes
based on the representations $ h(x_t) $. For the actual classification, however, a
global video representation $ \mathcal{H}(\vec{x}) $ is used.
In our approach, on the contrary, the codebook is optimized to distinguish the classes based on the
final representation $ \mathcal{H}(\vec{x}) $ directly rather than on an intermediate quantity
$ h(x_t) $.

\subsection{Equivalence Results}

There is a close relation between single-layer neural networks and Gaussian models~\cite{macherey2003comparative,heigold2007equivalence}.
We consider the special case of kMeans here. Following the derivation in the previous section,
the kMeans model can be transformed into a single-layer neural network.
For the other direction, however, the constraint that the bias components are inner products
of the weight matrix rows (see Equations \eqref{nnWeights} and \eqref{nnBias}) is not met when
optimizing the neural network parameters.
In fact, the single-layer neural network is equivalent to a kMeans model with non-uniform visual word priors $ p(v_m) $.
While the transformation from a kMeans model to a neural network is defined by Equations \eqref{nnWeights}
and \eqref{nnBias}, the transformation from the neural network model to a kMeans model is given by
\begin{align}
    v_m &= (\mat{W}_{1,m} \dots \mat{W}_{D,m})^\intercal, \\
    p_{\mathrm{NN}}(v_m) &= \frac{\exp{(b_m + \frac{1}{2}v_m^\intercal v_m)}}
                  {\sum_{\tilde m} \exp{(b_{\tilde m} + \frac{1}{2}v_{\tilde m}^\intercal v_{\tilde m})}}. \label{nnPrior}
\end{align}

\subsection{Encoding Kernels in the Neural Network}
\label{sec:featureMapLayer}

So far, the recurrent neural network is capable of computing bag-of-words like
histograms that are then used in a support vector machine in combination with a kernel. In this
section, we show how to incorporate the kernel itself into the neural network.

Consider the histograms $ h_1 $ and $ h_2 $ of two input sequences $ \vec{x} $
and $ \vec{y} $,
\begin{align}
    h_1 = \mathcal{H}(\vec{x}), \quad h_2 = \mathcal{H}(\vec{y}) \in \IR^M.
\end{align}
A kernel $ \mathcal{K}(h_1,h_2) $ is defined as the inner product
\begin{align}
    \mathcal{K}(h_1,h_2) = \langle \Psi(h_1), \Psi(h_2) \rangle,
\end{align}
where $ \Psi $ is the feature map inducing the kernel. In general, it is difficult
to find an explicit formulation of $ \Psi $. In~\cite{vedaldi2012efficient},
Vedaldi and Zisserman provide an explicit (approximate) representation for additive
homogeneous kernels. 
A kernel is called \textit{additive} if
\begin{align}
    \mathcal{K}(h_1, h_2) = \sum_{m=1}^M k(h_{1,m},h_{2,m}),
\end{align}
where $ k: \IR_0^+ \times \IR_0^+ \mapsto \IR_0^+ $ is a kernel induced by a
feature map $ \psi $. $ k $ is \textit{homogeneous} if
\begin{align}
    k(\alpha h_{1,m}, \alpha h_{2,m}) = \alpha k(h_{1,m},h_{2,m}).
\end{align}
According to~\cite{vedaldi2012efficient}, the feature map $ \psi $ for such kernels is
approximated by
\begin{align}
    [ \psi(x) ]_j =
        \begin{cases}
            \sqrt{\kappa (0)xL} & j = 0, \\
            \sqrt{2\kappa (\frac{j+1}{2}L)xL} \cos(\frac{j+1}{2}L \log x) & j \text{ odd}, \\
            \sqrt{2\kappa (\frac{j}{2}L)xL} \sin(\frac{j}{2}L \log x) & j \text{ even},
        \end{cases}
    \label{approximateFeatureMap}
\end{align}
where $ \kappa $ is a function dependent on the kernel, $ L $ is a sampling period,
and $ 0 \leq j \leq 2n $ defines the number of samples, see~\cite{vedaldi2012efficient}
for details.

\begin{figure}[t]
\begin{center}
    \includegraphics{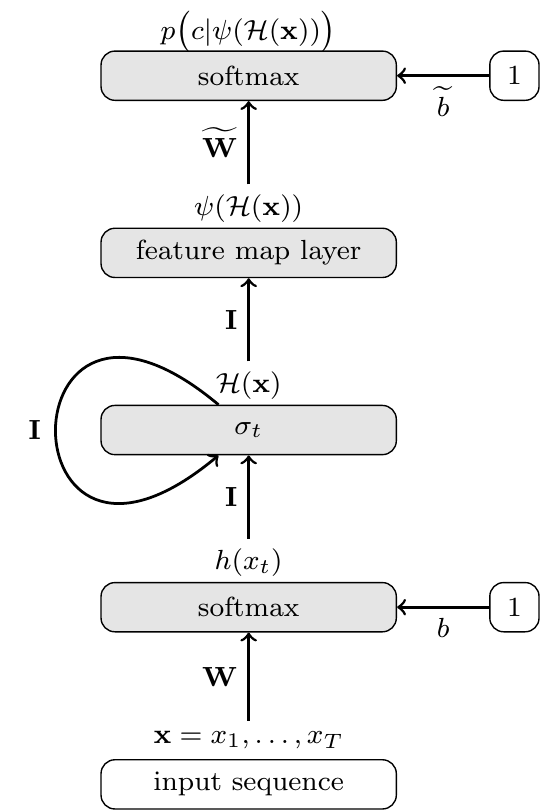}
\end{center}
\caption{Neural network encoding the bag-of-words model and a kernel via its feature map.}
\label{fig:svnNN}
\end{figure}
The function $ [ \psi(x) ]_j $ is continuously differentiable on the non-negative real
numbers and the derivative is given by
\begin{align}
    \frac{\partial [ \psi(x) ]_j}{\partial x} =
        \begin{cases}
            [\psi(x)]_0 \gamma(x) & j = 0, \\
            ([\psi(x)]_{j+1}(j+1)L + [\psi(x)]_j) \gamma(x) & j \text{ odd}, \\
            ([\psi(x)]_{j-1}jL + [\psi(x)]_j) \gamma(x) & j \text{ even},
        \end{cases}
    \label{approximateFeatureMapDerivative}
\end{align}
where
\begin{align}
    \gamma(x) = \frac{\kappa(0)L}{2[\psi(x)]_0^2}.
\end{align}
Since we apply kernels to histograms, the input to a feature map is always
non-negative in our case. Hence, Equation \eqref{approximateFeatureMap} can be
implemented as a layer in a neural network. Adding such a feature map layer between
the recurrent layer and the softmax output allows to represent the bag-of-words
pipeline including support vector machine and kernel computations completely in a single neural network,
\cf Figure \ref{fig:svnNN}.

To illustrate that this modification of the neural network is in fact sufficient
to model a support vector machine with a non-linear kernel, consider a simple two
class problem. The classification rule for the support vector machine is then
\begin{align}
    r_{\text{SVM}}(\mathcal{H}(\mat{x})) = \mathrm{sgn} \Big( \sum_{i=1}^I \alpha_i y_i \mathcal{K}(\mathcal{H}(\vec{x}_i),\mathcal{H}(\vec{x})) + b \Big)
\end{align}
with $ I $ support vectors and coefficients $ \alpha_i $ as well as labels $ y_i \in \{-1,1\} $.
Defining
\begin{align}
    \vec{w}_c = \sum_{i=1}^I \alpha_i y_i \Psi(\mathcal{H}(\vec{x}_i))
\end{align}
allows to simplify the decision rule to
\begin{align}
    r_{\text{SVM}}(\mathcal{H}(\mat{x})) = \mathrm{sgn} \big( \langle \vec{w}_c, \Psi(\mathcal{H}(\vec{x})) \rangle + b \big).
\end{align}
As can be seen from this equation, the decision rule is an inner product of a weight vector
and the feature map $ \psi(\mathcal{H}(\vec{x})) $. This is the same operation that is
performed in the neural network, apart from the $ \softmax $ output layer. This, however,
does not affect the maximizing argument, so the decision of the support vector machine
and the decision of the neural network are the same if the same weights and bias are
used. Still, in contrast to the support vector machine, the neural network is trained
according to the cross-entropy criterion using unconstrained optimization.
So in practice, the neural network model usually differs from the model obtained with a support
vector machine.

Note that the approximate feature map increases the dimension and, thus, also the
number of parameters, depending on the number of samples. If the histograms are
of dimension $ M $, the output of the feature map layer is of dimension $ M \cdot (2n+1) $.
In practice, however, $ n = 2 $ already works well~\cite{vedaldi2012efficient}.

\subsection{Implementation Details}

Neural networks are usually optimized using gradient based methods such as
stochastic gradient descent (SGD) or Resilient Propagation (RProp;~\cite{riedmiller1993direct}).
The gradient of a recurrent neural network can be efficiently computed using
backpropagation through time (BPTT;~\cite{werbos1990backpropagation}).
In a forward pass, the activation $ y_{l,t} $ of each layer $ l $ is computed
for all timeframes $ t $. The corresponding error signals $ e_{l,t} $ are then
computed in a backward pass through all layers and timeframes. Thereby, it is
necessary to keep all $ y_{l,t} $ in memory during backpropagation. For long
input sequences, this may be prohibitive since neural networks are usually
optimized on a GPU that has only a few GB of memory.

Our model is a specific kind of neural network with only a single recurrent
connection that has the unit matrix as weights. Two special properties emerge
from this structure. First, the order in which the input sequence $ x_1,\dots,x_T $
is presented to the network does not affect the output. Second, the error signals
of the recurrent layer are the same for each timeframe, \ie
\begin{align}
    e_{\mathrm{rec},T} = e_{\mathrm{rec},t} \quad \text{ for } t = 1,\dots,T.
\end{align}
Thus, it is sufficient to store $ e_{\mathrm{rec},T} $ once and then process each feature
vector in another pass through the network using standard error backpropagation~\cite{rumelhart1988learning},
which requires much less memory.
The training process is illustrated in Algorithm~\ref{algo1}. We indicate the first $ \mathrm{softmax} $
layer, the recurrent layer, the feature map, and the output layer as \textit{sft, rec, map,} and
\textit{out}, respectively.
Further, $ \mathbb{I}_c $ is used to indicate the unit vector with a one in its $ c $-th component
and $ \mathbf{J}_f(a) $ is the Jacobi matrix of a function $ f $ at $ a $. Given an input sequence $ \mathbf{x} $,
the activations of the output layer are computed in a forward pass. In the following backward pass,
the error signals for each timeframe are used to accumulate the gradients. Note that the algorithm
is a special case of BPTT. Due to the trivial recurrent connection, the activations and error signals
at times $ t < T $ do not need to be stored once the error signal $ e_{\mathrm{rec},T} $ is computed.

\begin{algorithm}[t]
    \caption{Memory efficient gradient computation}
    \label{algo1}
    \fontsize{10}{15}\selectfont
    \begin{algorithmic}[1]
        \State \textbf{input:} feature sequence $ \mathbf{x} $
        \State initialize $ \Delta \mathbf{W}, \Delta \mathbf{\widetilde W}, \Delta b, \Delta \tilde b $ with zero
        \Procedure{forward-pass}{}
            \For{$ t = 1,\dots,T $}
                \State $ y_{\mathrm{sft},t} \leftarrow \mathrm{softmax}(\mathbf{W}^\intercal x_t + b) $
                \State $ y_{\mathrm{rec},t} \leftarrow \sigma_t(y_{\mathrm{sft},t} + y_{\mathrm{rec},t-1}) $
                \State delete $ y_{\mathrm{sft},t} $ and $ y_{\mathrm{rec},t-1} $
            \EndFor
            \State $ y_{\mathrm{map},T} \leftarrow \psi(y_{\mathrm{rec},T}) $
            \State $ y_{\mathrm{out},T} \leftarrow \mathrm{softmax}(\mathbf{\widetilde W}^\intercal y_{\mathrm{map},T} + \tilde b) $
        \EndProcedure
        \Procedure{backward-pass}{}
            \State $ e_{\mathrm{out},T} \leftarrow y_{\mathrm{out},T} - \mathbb{I}_c $
            \State $ e_{\mathrm{map},T} \leftarrow \mathbf{J}_{\psi}(y_{\mathrm{rec},T}) \cdot \mathbf{\widetilde W} \cdot e_{\mathrm{out},T} $
            \State $ e_{\mathrm{rec},T} \leftarrow \mathbf{J}_{\sigma_T}(y_{\mathrm{sft},T}) \cdot e_{\mathrm{map},T} $
            \For{$ t = 1,\dots,T $}
                \State $ e_{\mathrm{sft},t} \leftarrow \mathbf{J}_{\mathrm{softmax}}(\mathbf{W}^\intercal x_t + b) \cdot e_{\mathrm{rec},T} $
                \State $ \Delta \mathbf{W} \leftarrow \Delta \mathbf{W} + x_t \cdot e_{\mathrm{sft},t}^\intercal $
                \State $ \Delta b \leftarrow \Delta b + e_{\mathrm{sft},t} $
            \EndFor
            \State $ \Delta \mathbf{\widetilde W} \leftarrow y_{\mathrm{map},T} \cdot e_{\mathrm{out},T}^\intercal $
            \State $ \Delta \tilde b \leftarrow e_{\mathrm{out},T}^\intercal $
       \EndProcedure
       \Return $ \Delta \mathbf{W}, \Delta \mathbf{\widetilde W}, \Delta b, \Delta \tilde b $
    \end{algorithmic}
\end{algorithm}



\section{Experimental Setup}
\label{sec:setup}

\subsection{Feature Extraction}

We extract improved dense trajectories as described in~\cite{wang2013action},
resulting in five descriptors with an overall number of 426 features per
trajectory, and apply z-score normalization to the data.
We distinguish between two kinds of features: concatenated and separated
descriptors. For the first, all $ 426 $ components of a trajectory are
treated as one feature vector. For the latter, the dense trajectories are
split into their five feature types \textit{Traj}, \textit{HOG}, \textit{HOF},
and two motion boundary histograms \textit{MBHX} and \textit{MHBY} in $ x $-
and $ y $-direction.

\subsection{kMeans Baseline}

For the baseline, we follow the approach of~\cite{wang2013dense}:
kMeans is run eight times on a randomly sampled subset of $ 100,000 $
trajectories. The result with lowest sum of squared distances is used as
visual vocabulary. For concatenated descriptors, a histogram of $ 4,000 $
visual words is created based on the $ 426 $-dimensional dense trajectories.
In case of separate descriptors, a visual vocabulary with $ 4,000 $ visual words
is computed for each descriptor type separately. The resulting five histograms are
combined with a multichannel RBF-$ \chi^2 $ kernel as proposed in~\cite{wang2013dense},
\begin{align}
    \mathcal{K}(i,j) = \exp{ \bigg( -\frac{1}{5}\sum_{c=1}^5 \frac{D(\mathcal{H}(\vec{x}_i^c), \mathcal{H}(\vec{x}_j^c))}{A_c} \bigg) },
    \label{multichannelChiSquare}
\end{align}
where $ \vec{x}_i^c $ is the $ c $-th descriptor type of the $ i $-th video,
$ D(\cdot,\cdot) $ is the $ \chi^2 $-distance between two histograms, and $ A_c $
is the mean distance between all histograms for descriptor $ c $ in the training
set. For concatenated features, the kernel is used with a single channel only.
As classifier, we train a \textit{one-against-rest} support vector machine using
LIBSVM~\cite{chang2011libsvm}.

\subsection{Neural Network Setup}

When training neural networks, the trajectories of each video are uniformly
subsampled to reduce the total amount of trajectories. The network is trained
according to the cross-entropy criterion, which maximizes the likelihood of the
posterior probabilities. We use RProp as optimization algorithm and iterate until
the objective function does not improve further. If the neural network output
is not directly used for classification, \ie if a support vector machine is
used to classify the histograms generated by the neural network, overfitting
is not a critical issue. Thus, strategies like regularization or dropout do not
need to be applied. Furthermore, we could not investigate any advantages when
initializing with a kMeans model. Normalization, in contrast, is crucial. If the
network input is not normalized, the training of the neural network is highly
sensitive to the learning rate and RProp even fails to converge. For consistency
with the kMeans baseline, the number of units in the first $ \softmax $ layer and
the recurrent layer that computes the histograms is also set to $ 4,000 $.

\subsection{Datasets}

For the evaluation of our method, we use four action recognition benchmarks,
two of which are of medium and two of large scale.

With $ 783 $ action clips from $ 16 $ classes, the \textbf{Olympic Sports}
dataset \cite{niebles2010modeling} is the smallest among the four benchmarks.
The videos show athletes performing Olympic disciplines and are several seconds
long. We use the train/test split suggested in \cite{niebles2010modeling}, which
partitions the dataset into $ 649 $ training videos and $ 134 $ test instances.
After extracting improved dense trajectories, the training set comprises about
$ 40 $ million trajectories and the test set $ 7.8 $ million. For evaluation,
mean average precision is reported.

\textbf{HMDB-51} \cite{kuehne11hmdb} is a large scale action recognition benchmark
containing $ 6,849 $ clips of $ 51 $ different classes. The clips are collected
from public databases and movies. In contrast to Olympic Sports, the clips in
HMDB-51 are usually only a few seconds long.
The dataset provides at least $ 101 $ instances of each action class and the authors
propose a three-fold cross validation. All splits are of comparable size and after
feature extraction, there are $ 42 $ million trajectories in the training set and
$ 18 $ million in the test set. We report average accuracy over the splits.

In order to validate the applicability of our method to datasets of different sizes,
we also conduct experiments on a subset of HMDB-51.
\textbf{J-HMDB}~\cite{jhuang2013jhmdb} comprises a subset of $ 928 $ videos from $ 21 $
action classes. With a range from $ 15 $ to $ 40 $ frames, the clips are rather short.
We follow the protocol of \cite{jhuang2013jhmdb} and use three splits. Each split
partitions the videos into training and test set with an approximate ratio of
$ 70:30 $. For the training set, about two million trajectories are extracted for
each split. For the test set, the amount of extracted trajectories ranges between
$ 700,000 $ and $ 800,000 $ depending on the split. Even though the number of clips
in the dataset is in the same order as for Olympic Sports, it is clearly the smallest
dataset in our evaluation in terms of extracted trajectories. As for HMDB-51, average
accuracy over the three splits is reported.

The largest benchmark we use is the \textbf{UCF101} dataset \cite{soomro2012ucf101}.
Comprising $ 13,320 $ video clips from a set of $ 101 $ different action classes,
it is about twice as large as HMDB-51. The dataset contains videos from five
major categories (sports, human-human-interaction, playing musical instruments,
body-motion only, and human-object interaction) and has been collected from
Youtube. Again, we follow the protocol of \cite{soomro2012ucf101} and use the
suggested three splits. Each split partitions the data in roughly $ 9,500 $ training
clips and $ 3,700 $ test clips, corresponding to $ 230 $ million improved trajectories
for the training set and $ 90 $ million for the test set, respectively.
Again, we report average accuracy over the three splits.

For Olympic Sports and HMDB-51, we use the human bounding boxes provided by Wang
and Schmid.\footnote{http://lear.inrialpes.fr/people/wang/improved\_trajectories}
For the other datasets, improved dense trajectories are extracted without human bounding
boxes.



\section{Experimental Results}
\label{sec:experiments}

In this section, we evaluate our method empirically. In a first step, the impact
of the prior and the difference in the histograms generated by a standard kMeans
model and our neural network are analyzed. Then, the effect of subsampling the
dense trajectories is investigated. We show that even with a small number of
trajectories, satisfying results can be obtained while accelerating the training
time by up to two orders of magnitude. Moreover, we compare our model
to the standard pipeline for action recognition from~\cite{wang2013dense, wang2013action}
and show that the method is not only bound to action recognition but is also
competitive to some well known sparse coding methods in image classification.
We also evaluate the effect approximating the kernel directly within the neural
network as proposed in Section \ref{sec:featureMapLayer}, followed by a comparison
to the current state-of-the-art in action recognition.

\subsection{Evaluation of the Neural Network Model}

\begin{figure*}[t]
\begin{center}
    \small
    \subfloat[kMeans prior $ p_{\mathrm{KM}}(v_m) $]{
        \includegraphics{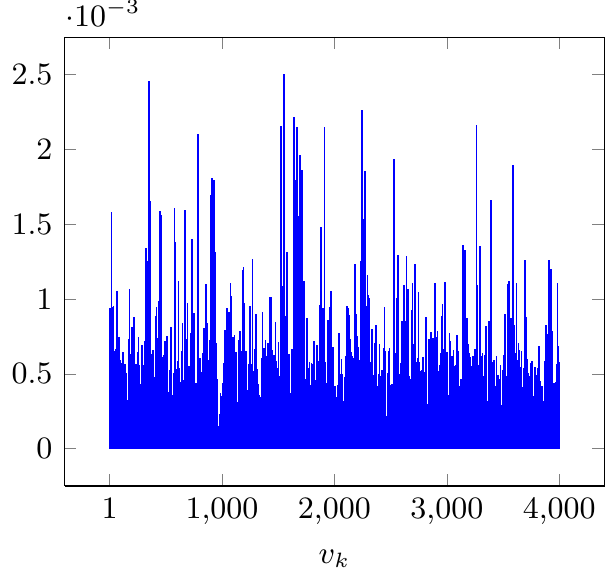}
        \label{fig:priorKM}
    }\hspace{2cm}
    \subfloat[neural network prior $ p_{\mathrm{NN}}(v_m) $]{
        \includegraphics{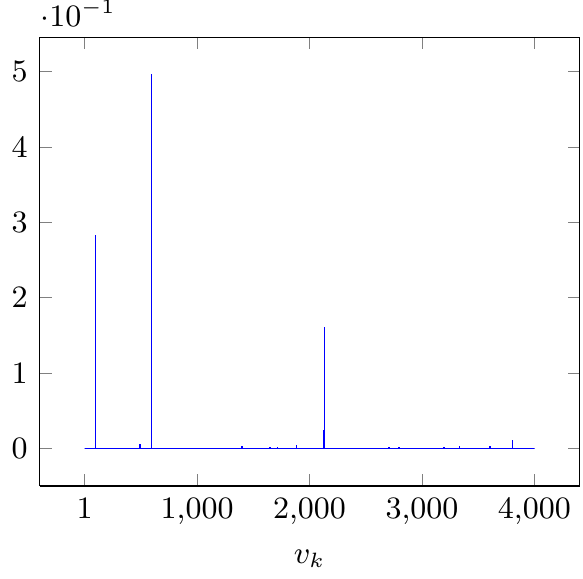}
        \label{fig:priorNN}
    }
\end{center}
\caption{Comparison of the visual word prior induced by kMeans and the prior
         obtained with the neural network on the Olympic Sports dataset.
         Note the different scales on the $ y $-axes.}
\end{figure*}
%
\begin{figure*}[!]
\begin{center}
    \subfloat[Example frame (left), kMeans histogram (middle), and neural network histogram (right) for the video \textit{high\_jump/010.avi}.]{
        \raisebox{27pt}{\includegraphics[width=0.27\textwidth]{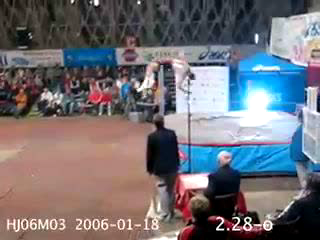}}
        \hspace{0.5cm}
        \includegraphics{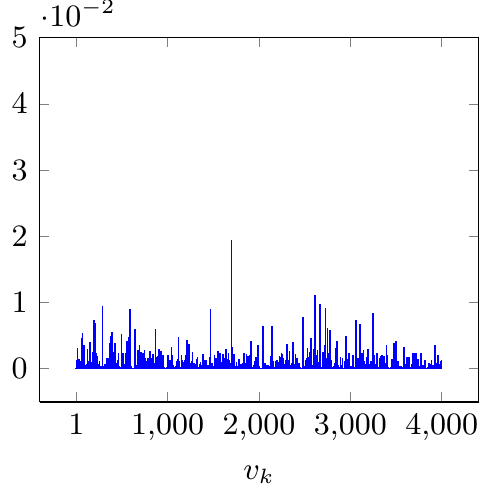}
        \hspace{0.5cm}
        \includegraphics{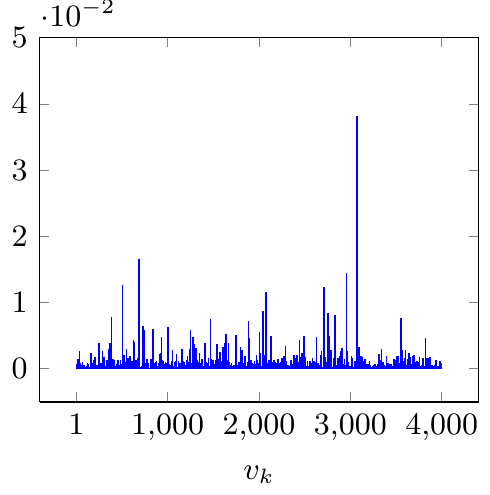}
    } \\
    \subfloat[Example frame (left), kMeans histogram (middle), and neural network histogram (right) for the video \textit{high\_jump/011.avi}.]{
        \raisebox{27pt}{\includegraphics[width=0.27\textwidth]{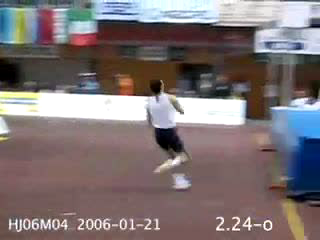}}
        \hspace{0.5cm}
        \includegraphics{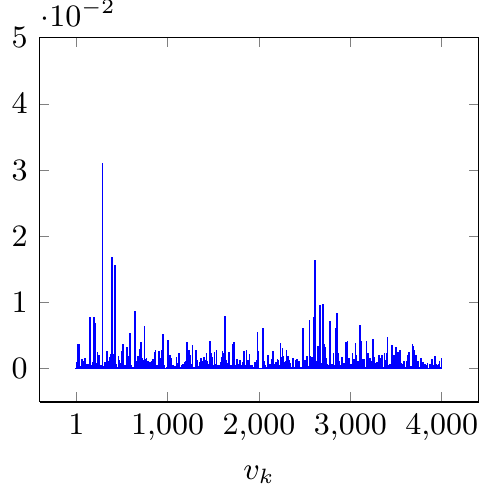}
        \hspace{0.5cm}
        \includegraphics{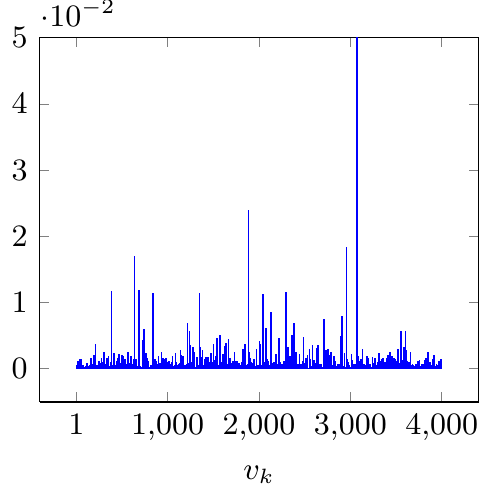}
    } \\
    \subfloat[Example frame (left), kMeans histogram (middle), and neural network histogram (right) for the video \textit{pole\_vault/001.avi}.]{
        \raisebox{27pt}{\includegraphics[width=0.27\textwidth]{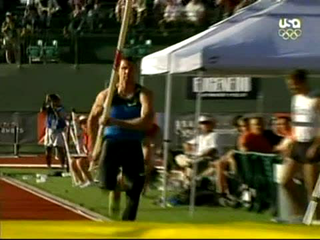}}
        \hspace{0.5cm}
        \includegraphics{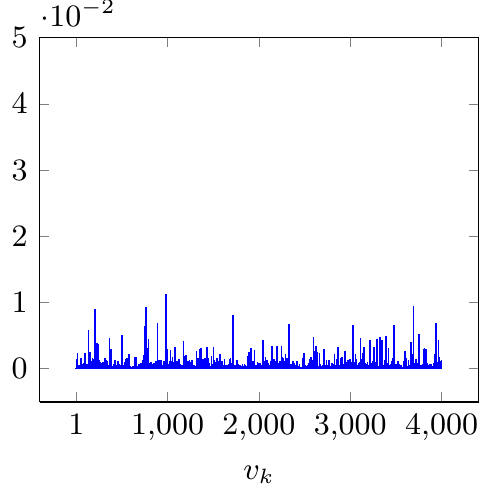}
        \hspace{0.5cm}
        \includegraphics{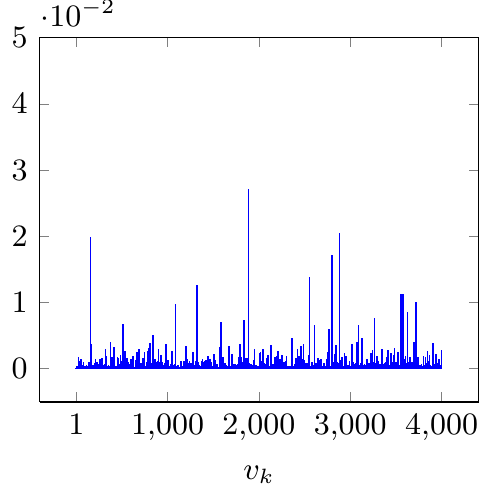}
    } \\
    \subfloat[Example frame (left), kMeans histogram (middle), and neural network histogram (right) for the video \textit{pole\_vault/005.avi}.]{
        \raisebox{27pt}{\includegraphics[width=0.27\textwidth]{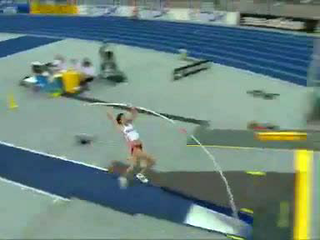}}
        \hspace{0.5cm}
        \includegraphics{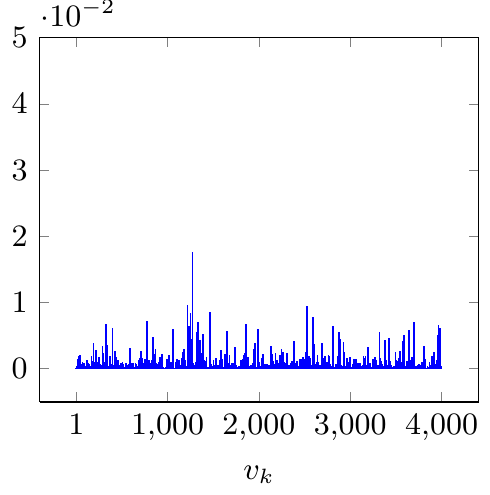}
        \hspace{0.5cm}
        \includegraphics{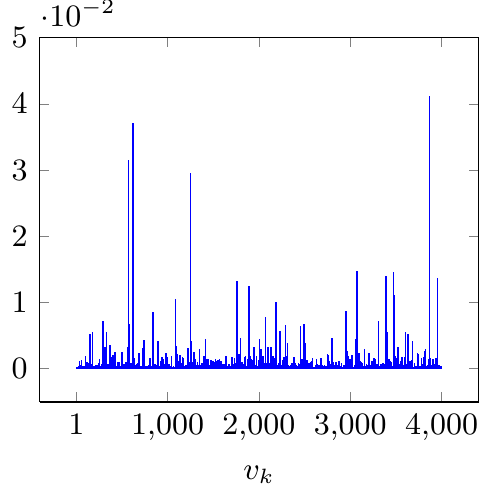}
    }
\end{center}
\caption{Comparison of the histograms generated by the kMeans model and the neural
         network model for four videos of two different classes from Olympic Sports.}
\label{fig:histograms}
\end{figure*}
We evaluate the neural network model by comparing its performance on Olympic
Sports to the performance of the kMeans based bag-of-words model. Moreover,
we analyze the difference between the histograms generated by the neural network
and those generated by the kMeans approach. For neural network training, the
number of trajectories per video is limited to $ 5,000 $ using uniform subsampling.

As mentioned in Section \ref{sec:technical}, the neural network implicitly models
a visual word prior $ p_{\mathrm{NN}}(v_m) $. Hence, we compare to an additional version of kMeans in
which we compute the posterior distribution $ p_{\mathrm{KM}}(v_m|x) $ with a
non-uniform prior $ p_{\mathrm{KM}}(v_m) $. The prior is modeled as relative
frequencies of the visual words. Note that the resulting model is equivalent to
the neural network model and only differs in the way the parameters are estimated.

\begin{table}
\begin{center}
    \newcolumntype{R}{>{\raggedleft\arraybackslash}X}
    \begin{tabularx}{0.45\textwidth}{lXlXr}
        \toprule
                        & & \multicolumn{3}{c}{assignment} \\
                            \cmidrule(lr){3-5}
                        & & soft       & & hard \\
        \midrule
        kMeans          & & $ 84.1\% $ & & $ 84.0\% $ \\
        kMeans + prior  & & $ 83.7\% $ & & $ 84.0\% $ \\
        neural network  & & $ 86.7\% $ & & $ 86.3\% $ \\
        \bottomrule
    \end{tabularx}
\end{center}
\caption{Comparison of three bag-of-words models: kMeans, kMeans with prior, and
         the neural network (Figure \ref{fig:nn}). For each model, the histograms are once computed using
         soft assignment, \ie the posterior distribution $ p(v_m|x) $ is used directly,
         and once using hard assignment, \ie shifting all probability mass to the most likely visual
         word.}
\label{tab:maxApprox}
\end{table}

When computing histograms of visual words, usually hard assignment is used
as in $ \hat p_{\mathrm{KM}}(v_m|x) $ in Equation \eqref{probKMeans}. This may be
natural for kMeans since during the generation of the visual vocabulary, each
observation only contributes to its nearest visual word. For the neural network
model, on the contrary, hard assignment can not be used during training since
differentiability is required. Thus, it is natural to use the posterior distribution
$ p_{\mathrm{NN}}(v_m|x) $ directly for the histogram computation. However, in Table
\ref{tab:maxApprox} it can be seen that there is no significant difference between
soft and hard assignment for either of the three methods.

Furthermore, Table \ref{tab:maxApprox} reveals that the neural network is more
than $ 2\% $ better than the kMeans model. However, the improvement can not be
explained by the additional degrees of freedom, as the kMeans model with non-uniform
prior does not improve compared to the original kMeans model. Discriminative
training allows the neural network to generate visual words that discriminate well
between the classes. KMeans, in contrast, only generates visual words that represent
the observation space well regardless of the class labels. We validate this major
difference between both methods by a comparison of the visual word priors. In Figure
\ref{fig:priorKM}, the prior induced by kMeans, $ p_{\mathrm{KM}}(v_m) $, is illustrated.
The probability for all visual words is within the same order of magnitude. The neural
network prior $ p_{\mathrm{NN}}(v_m) $ is computed by means of Equation \eqref{nnPrior}
and depicted in Figure \ref{fig:priorNN}. Almost all probability mass is distributed
over three visual words. All other visual words are extremely rare, making their
occurrence in a histogram a very discriminative feature.

In Figure \ref{fig:histograms}, the histograms generated by kMeans and the neural network
are compared for four videos from two different classes, \textit{high\_jump} and \textit{pole\_vault}.
The first column shows an example frame of the video. The second and third column show the histograms
generated by the kMeans model and by the neural network, respectively.
The histograms generated by the neural network are sharper than those generated by the kMeans model.
The two peaks in the neural network histograms of the \textit{high\_jump} videos
at visual word index $ 3,000$ are a pattern that occurs in multiple histograms of this class.
Similar patterns can also be observed for neural network histograms of other classes
but usually not for kMeans based histograms, although the example videos from \textit{high\_jump}
are very similar in appearance.
If the appearance of the videos undergoes stronger variations as in the \textit{pole\_vault}
class, these reoccurring patterns are neither observable for the kMeans based histograms
nor for the neural network histograms. Still, the latter have clearer peaks that
lead to larger differences between histograms of different videos, raising potential for a
better discrimination. This confirms the preceding evaluation of the results in Table
\ref{tab:maxApprox}.

During training of the neural network, a class posterior distribution $ p(c|\mathcal{H}(\vec{x})) $
is modeled. Using this model instead of the SVM with RBF-$ \chi^2 $ kernel for
classification is worse than the baseline. For concatenated descriptors, the result
on Olympic Sports is $ 82.3\% $. Regularization, dropout, and adding additional
layers did not yield any improvement. However, considering that the model for
$ p(c|\mathcal{H}(\vec{x})) $ is only a linear classifier on the histograms
(\cf Section \ref{sec:technical}), the result is remarkable as the kMeans baseline
with a linear support vector machine reaches only $ 69.6\% $.

\subsection{Effect of Feature Subsampling}
\label{sec:subsampling}

\begin{figure*}[t]
\begin{center}
    \small
    \subfloat{
        \includegraphics{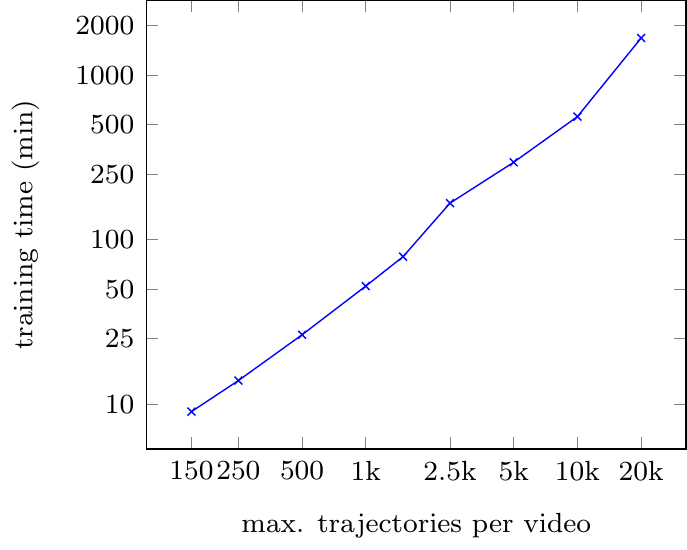}
    }\hfill
    \subfloat{
        \includegraphics{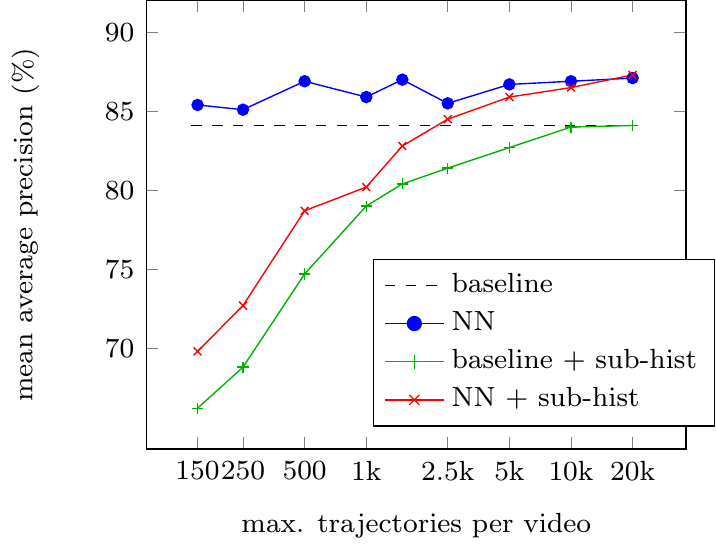}
     }
\end{center}
\caption{Effect of feature subsampling on the training time and the performance evaluated on Olympic Sports.
         Left: Time required to train the neural network.
         Right: Performance of the neural network (NN) based bag-of-words. For the blue curve, the number of
         trajectories has only been limited for neural network training. For the red and green curve, it has
         also been limited for the histogram computation on the training and test set (sub-hist).}
\label{fig:reducedTraj}
\end{figure*}

We evaluate the runtime and accuracy of our method when reducing the number
of trajectories per video on Olympic Sports. The networks are trained on a GeForce
GTX $ 780 $ with $ 3 $GB memory. We limit the number of trajectories per video
to values from $ 150 $ to $ 20,000 $ via uniform subsampling. This corresponds to
an overall number of trajectories between $ 100,000 $ and $ 12 $ million.
In Figure \ref{fig:reducedTraj} (left), the runtime is shown.
When sampling $ 150 $ trajectories per video, which corresponds to $ 100,000 $
dense trajectories overall, both, our GPU implementation of kMeans and the
neural network training, run for nine minutes. The training times scales linearly
with the number of trajectories per video.
In Figure \ref{fig:reducedTraj} (right), the performance of the system with limited number
of trajectories is illustrated. For the blue curve, the number of trajectories has
only been limited for neural network training, but the histograms are computed on
all extracted trajectories. The performance of the neural network models is always
above the baseline (dashed line). The curve stabilizes around $ 5,000 $
trajectories per video, suggesting that this number is sufficient for the neural
network based bag-of-words. Note that for kMeans, we observed only small
fluctuations around one percent when changing the number of clustered trajectories,
the subsampling strategy, or the initialization. 

If the histograms for the training and test set are also computed on the limited set of trajectories (red and green curve),
the performance of the systems is much more sensitive to the number of subsampled
trajectories. However, when more than $ 5,000 $ trajectories per video (overall
$ 3.2 $ million trajectories) are used for the neural network histogram computation, the difference
to taking all extracted trajectories is small. Hence, it is possible to achieve
satisfying results with only $ 8\% $ of the originally extracted trajectories,
allowing to accelerate both, the histogram computation and the feature extraction
itself. A similar reduction is also possible with kMeans, but the loss in accuracy
is higher, \cf Figure \ref{fig:reducedTraj}.

\subsection{Comparison to Other Neural Network Architectures}

\begin{table}
\begin{center}
    \begin{tabularx}{0.45\textwidth}{lXr}
        \toprule
        \multicolumn{2}{l}{Method}  & mAP \\
        \midrule
        \multicolumn{2}{l}{\textit{Training on frame-level}} \\
        (a) & ours w/o recurrency                       & $ 62.3\% $ \\
        (b) & fine-tuned ImageNet CNN                   & $ 76.6\% $ \\
        \multicolumn{2}{l}{\textit{Training on video-level}} \\
        (c) & simple RNN                                & $ 18.3\% $ \\
        (d) & RNN with GRUs                             & $ 13.3\% $ \\
        (e) & attention-based RNN                       & $ 28.8\% $ \\
        (f) & ours                                      & $ 86.7\% $ \\
        \bottomrule
    \end{tabularx}
\end{center}
\caption{Comparison of different neural network architectures on Olympic Sports. Results are reported as mean average precision.}
\label{tab:otherNnArchitectures}
\end{table}

In this section, we compare our proposed model to other neural network architectures.
Our analysis is twofold: firstly, we show the importance of video-level training,
secondly, we prove that the bag-of-words equivalent architecture is crucial and other
recurrent neural networks fail to obtain good results.

Addressing the first point, we remove the recurrent connection from our model. The result
is a standard feed-forward network. For training, we assign to each input frame the class label
of the respective video. For inference, we cut off the last softmax layer and compute a video
representation by average pooling over the network output of each frame. Following the same
protocol as for our recurrent network, the resulting representation is transformed with a
multi-channel $ \chi^2 $ kernel and classified using a support vector machine.
In this setup, each input frame is treated separately and temporal context is not considered
during codebook training. The resulting loss of accuracy is significant: the performance
drops from $ 86.7\% $ to $ 62.3\% $, \cf Table \ref{tab:otherNnArchitectures} (a) and (f).
As another example for frame-level training, we fine-tuned the $ 51 $-layer deep residual CNN
of the $ 2015 $ ImageNet challenge winning submission \cite{he2015deep} on Olympic Sports.
In order to get class posterior probabilities for a complete video, we applied average
pooling to the framewise posteriors. Still, this state-of-the-art CNN architecture struggles
to achieve competitive results.

Addressing the second point, we replace the bag-of-words equivalent architecture by other
recurrent network architectures. \textit{Simple RNN} (c) is a neural network with a single
recurrent hidden layer and sigmoid activations. \textit{RNN with GRUs} (d) has the same
structure but the sigmoid units are replaced by gated recurrent units \cite{cho2014learning}.
Those units allow the network to decide when to discard past frames and when to update the
recurrent activations based on forget and update gates.
Not surprisingly, the performance of both recurrent architectures is poor. Recurrent neural
networks tend to forget about past inputs exponentially fast, so particularly for long
sequences as we have in our setup, only the end of the video is actually observed and the
crucial parts at the beginning and in the middle of the video are not considered.
\textit{Attention-based RNNs} \cite{bahdanau2015neural} aim to solve this problem. An
attention layer learns weights for each timeframe of a recurrent layer and accumulates
the recurrent layer activations according to this weighting. This way, information is
captured along the complete sequence. Unfortunately, these models have a large number of
parameters and are highly sensitive to overfitting, especially in case of the huge
intra-class variations in video data. Consequently, the classification result is better
than for traditional recurrent neural networks but still not competitive to state-of-the-art
methods, \cf Table \ref{tab:otherNnArchitectures} (e).

Summarizing, the results in Table \ref{tab:otherNnArchitectures} show that both the
specific architecture and the joint training of classifier and vocabulary on video-level
are crucial.

\subsection{Evaluation on Various Datasets}

\begin{table*}
\begin{center}
    \begin{tabularx}{\textwidth}{lXrrXrr}
        \toprule
        \multicolumn{2}{r}{Descriptors:} & \multicolumn{2}{c}{concatenated} & & \multicolumn{2}{c}{separate} \\
                                           \cmidrule(lr){3-4}                 \cmidrule(lr){6-7}
                       & & baseline   & neural network      & & baseline   & neural network      \\
        \midrule
        Olympic Sports & & $ 84.1\% $ & $ \mathbf{86.7\%} $ & & $ 84.4\% $ & $ \mathbf{85.9\%} $ \\
        J-HMDB         & & $ 56.6\% $ & $ \mathbf{57.6\%} $ & & $ 59.1\% $ & $ \mathbf{61.9\%} $ \\
        HMDB-51        & & $ 45.8\% $ & $ \mathbf{50.6\%} $ & & $ 52.2\% $ & $ \mathbf{54.0\%} $ \\
        UCF101         & & $ 67.8\% $ & $ \mathbf{73.3\%} $ & & $ 73.3\% $ & $ \mathbf{76.9\%} $ \\
        \bottomrule
    \end{tabularx}
\end{center}
\caption{Comparison of the kMeans baseline with the neural network model for concatenated
         and separate descriptors on four different datasets.}
\label{tab:finalResults}
\end{table*}

We evaluate our method on the four action recognition datasets Olympic
Sports, J-HMDB, HMDB-51, and UCF101. On J-HMDB, all extracted trajectories are
used for neural network training. On HMDB-51 and Olympic Sports, we limit the
number of trajectories per video to $ 5,000 $ as proposed in Section
\ref{sec:subsampling}. For terms of efficiency, we further reduce this number to
$ 2,500 $ for the largest dataset UCF101. We conduct the experiments for concatenated
descriptors, \ie we directly use a $ 426 $ dimensional feature vector for each trajectory,
and for separate descriptors as originally proposed in~\cite{wang2013dense}.
The results are shown in Table \ref{tab:finalResults}.

The neural network outperforms the baseline on all datasets. For the smaller
datasets J-HMDB and Olympic Sports that have only few classes, the improvement is
between $ 1\% $ and $ 2.6\% $ in case of concatenated descriptors. For the large
datasets, however, the baseline is outperformed by around $ 5\% $.
In case of separate descriptors, the improvement is smaller but still ranges from
$ 1.5\% $ to $ 3.6\% $.

Comparing the neural network with concatenated descriptors (second column of
Table \ref{tab:finalResults}) and the baseline with separate descriptors (third
column of Table \ref{tab:finalResults}) reveals that both systems achieve
similar accuracies. However, for the baseline with separate descriptors,
visual vocabularies and histograms have to be computed for each of the five
descriptors separately. For the neural network with concatenated descriptors,
in contrast, it is sufficient to train a single system.

\subsection{Application to Image Classification}

\begin{table*}
\begin{center}
    \begin{tabularx}{\textwidth}{X r r r}
        \toprule
        Method                                           & Codebook Size      & Caltech-101 & 15-Scenes  \\
        \midrule
        Hard assignment~\cite{lazebnik2006beyond}        & $ 200 $            & $ 64.6\% $  & $ 81.1\% $ \\
        Kernel Codebooks~\cite{van2010visual}            & $ 200 $            & $ 64.1\% $  & $ 76.7\% $ \\
        Soft assignment~\cite{liu2011defense}            & $ 1000 $           & $ 74.2\% $  & $ 82.7\% $ \\
        \midrule
        ScSPM~\cite{yang2009linear}                      & $ 1024 $           & $ 73.2\% $  & $ 80.3\% $ \\
        LLC~\cite{wang2010locality}                      & $ 2048 $           & $ 73.4\% $  & -          \\
        Multi-way local pooling~\cite{boureau2011ask}    & $ 1024 \times 65 $ & $ 77.8\% $  & $ 83.1\% $ \\
        \midrule
        Unsupervised SS-RBM~\cite{goh2012unsupervised}   & $ 1024 $           & $ 75.1\% $  & $ 84.1\% $ \\
        \textbf{Ours}                                    & $ \mathbf{1024} $  & $ \mathbf{74.5\%} $  & $ \mathbf{83.5\%} $ \\
        \bottomrule
    \end{tabularx}
\end{center}
\caption{Comparison of our method to other coding methods on Caltech-101 and 15-scenes.}
\label{tab:image}
\end{table*}

Although designed to meet some specific problems in action recognition, our method is applicable to image
datasets, too. 
We compare to existing sparse coding methods on two small image datasets, Caltech-101~\cite{fei2007learning}
and 15-scenes~\cite{fei2005bayesian}. Following the setup of~\cite{goh2012unsupervised}, we densely extract
SIFT features, compute spatial pyramids, and use a linear support vector machine for classification. Note that our method is not
particularly designed for such a setting since we do not train our encoding directly on the spatial pyramid
features that are finally used for classification. In contrast to the methods~\cite{yang2009linear,wang2010locality,boureau2011ask,goh2012unsupervised}, we do not introduce any sparsity
constraints.
Still, our method shows competitive results compared to several
other coding methods, see Table \ref{tab:image}. 

\subsection{Using Explicit Feature Maps}

In this section, we analyze the effect of explicit feature maps that allow to train
the complete system in a single network. We use the network architecture from
Figure \ref{fig:svnNN}.
Since the multichannel RBF-$ \chi^2 $ kernel from Equation \eqref{multichannelChiSquare}
is not an additive homogeneous kernel, we evaluate three different additive homogeneous
kernels here:
\textit{Hellinger's kernel} is defined as
\begin{align}
    k(h_{1,i},h_{2,i}) = \sqrt{h_{1,i} h_{2,i}}
\end{align}
and its underlying feature map has an exact closed form solution, $ \psi(x) = \sqrt{x} $.
It is notable that due to the non-negativity of the histograms that are
fed into the function, Hellinger's kernel is equivalent to the application
of power normalization that has been proposed for Fisher vectors in
\cite{perronnin2010improving}.

Moreover, we examine an additive homogeneous version of the $ \chi^2 $ kernel,
\begin{align}
    k(h_{1,i},h_{2,i}) = 2 \frac{h_{1,i}h_{2,i}}{h_{1,i} + h_{2,i}}.
\end{align}
In combination with a support vector machine, the kernel can be applied directly.
When incorporating it into the neural network, we use the approximation from
Equation \eqref{approximateFeatureMap} with $ \kappa(\lambda) = \mathrm{sech}(\pi \lambda) $.
Finally, the histogram intersection kernel
\begin{align}
    k(h_{1,i},h_{2,i}) = \min(h_{1,i},h_{2,i})
\end{align}
is approximated similarly with
\begin{align}
    \kappa(\lambda) = \frac{2}{\pi (1+4\lambda^2)}.
\end{align}
Detailed derivations for the $ \kappa $ functions can be found in~\cite{vedaldi2012efficient}.
As proposed in~\cite{vedaldi2012efficient}, we use $ n = 2 $ and $ L = 0.5 $ for both the $ \chi^2 $
kernel and the histogram intersection kernel. In case of concatenated descriptors,
this increases the amount of units from $ 4,000 $ in the histogram layer to $ 20,000 $
in the feature map layer.

In order to be able to train a single neural network not only for concatenated
descriptors but also for separate descriptors, each of the five descriptors is handled
in an independent channel within the neural network allowing to train five independent
visual vocabularies. After application of the feature map, the five channels are
concatenated to a single channel five times the size of each feature map. A $ \softmax $
layer is used for classification on top of the concatenated channel. As a consequence, the
number of units after application of the feature map is five times larger than for
concatenated descriptors, finally ending up with $ 100,000 $ units. This affects the
runtime for both, training and recognition, and can be particularly important for larger
datasets.

The application of Hellinger's kernel, the $ \chi^2 $ kernel, and the histogram intersection
kernel in combination with a support vector machine for concatenated descriptors is
straightforward. In case of separate descriptors, we combine the five histograms by
concatenation.

\begin{table*}
\begin{center}
    \begin{tabularx}{\textwidth}{l X r r r X r r r}
        \toprule
                                         & & \multicolumn{3}{c}{concatenated}    & & \multicolumn{3}{c}{separate} \\
                                           \cmidrule(lr){3-5}                                    \cmidrule(lr){7-9}
        \multicolumn{2}{c}{feature map:}   & Hellinger & $ \chi^2 $ & hist. int. & & Hellinger & $ \chi^2 $ & hist. int. \\
        \midrule
        \textit{Olympic Sports}          & & \phantom{Hellinger} & \phantom{Hellinger} & \phantom{Hellinger} & & \phantom{Hellinger} & \phantom{Hellinger} & \phantom{Hellinger} \\
        from scratch                     & & $ 75.5 $  & $ 84.1 $   & $ 82.9 $   & & $ 84.3 $  & $ 81.4 $   & $ 80.4 $ \\
        init linear                      & & $ 81.3 $  & $ 85.0 $   & $ 83.8 $   & & $ 83.7 $  & $ 83.0 $   & $ 83.0 $ \\
        retrain top                      & & $ 86.4 $  & $ 85.8 $   & $ 85.8 $   & & $ 78.5 $  & $ 83.5 $   & $ 84.1 $ \\
        \midrule
        \textit{J-HMDB} \\
        from scratch                     & & $ 53.2 $  & $ 52.8 $   & $ 52.4 $   & & $ 56.9 $  & $ 45.9 $   & $ 46.2 $ \\
        init linear                      & & $ 50.6 $  & $ 55.1 $   & $ 53.6 $   & & $ 57.9 $  & $ 57.2 $   & $ 61.4 $ \\
        retrain top                      & & $ 57.5 $  & $ 56.6 $   & $ 56.7 $   & & $ 56.9 $  & $ 60.9 $   & $ 60.5 $ \\
        \midrule
        \bottomrule
    \end{tabularx}
\end{center}
\caption{Different training strategies when using a feature map layer.}
\label{tab:scratchLinearTop}
\end{table*}
We start with an evaluation of three different neural network training approaches on
the two smaller datasets J-HMDB and Olympic Sports. Training \textit{from scratch} starts with
a random initialization of all parameters, \textit{init linear} initializes the weights $ \mat{W} $
and bias $ \vec{b} $ with those obtained from the network from Figure \ref{fig:nn} which is trained with
a linear $ \softmax $ classifier on top of the histogram layer.
For \textit{retrain top} we took the same initialization but kept it fixed during training
and only optimized the parameters of the softmax output layer.

The results are shown in Table \ref{tab:scratchLinearTop}.
For concatenated descriptors, retraining only the top layer leads to the best results
for each feature map on both datasets and training from scratch performs worst in most
cases. For separate descriptors, however, retrain top is not always best. Particularly
on Olympic Sports, Hellinger's kernel fails to achieve competitive results. Still, for
all other cases retrain top is either best or at least close to the best training strategy.
Since it is the fastest among the three strategies, we find it beneficial particularly
for larger datasets.

Based on these results, we stick to the \textit{retrain top} strategy for large
datasets and compare the results to the traditional bag-of-words model and the neural
network plus support vector machine model from Section \ref{sec:conversionIntoNN}.
Table \ref{tab:differentKernels} shows the results of each of the three approaches for
three different kernels and both concatenated and separate descriptors.
The fourth and eighth column contain the numbers from Table \ref{tab:finalResults} in
order to provide a comparison to the originally used multichannel RBF-$\chi^2 $ kernel.
Recall that this kernel is not homogeneous and additive, so it can not be modeled as
a neural network layer.

For concatenated descriptors, both neural network approaches usually outperform the kMeans
baseline. While the difference between the kMeans baseline (\textit{kMeans + SVM}) and the
neural network including the feature map layer (\textit{retrain top}) is rather small, a
huge improvement can be observed for the neural network based visual words in combination
with a support vector machine (\textit{neural network + SVM}). Especially on
the larger datasets HMDB-51 and UCF101, about $ 6\% $ improvement is achieved. This
is particularly remarkable as the performance of a traditional bag-of-words model with
separate descriptors is almost reached although the model with concatenated descriptors
is much simpler: Only a single visual vocabulary with $ 4,000 $ visual words is computed,
while separate descriptors have an own visual vocabulary for each descriptor type, resulting
in a five times larger representation.

Analyzing the results for separate descriptors, the neural network with a feature map
corresponding to Hellinger's kernel shows surprisingly bad results on all datasets.
One explanation is that the feature map is simply the square root of each histogram
entry, while for all other kernels the approximate feature map from Equation \eqref{approximateFeatureMap}
has to be used. The number of units in the feature map layer is increased in the latter
case, raising potential for a better discrimination.

Similar to the results of concatenated descriptors, the neural network combined with
a support vector machine achieves the best results, whereas the neural network with a
feature map layer yields results comparable to the traditional bag-of-words model.
On all datasets, the best results have been obtained with separate descriptors and a
neural network plus support vector machine using a $ \chi^2 $ kernel, see second row,
sixth column for each dataset. Still, including a feature map layer also shows
competitive results in most cases while considerably reducing the runtime. Especially
on UCF101, the largest of the datasets, computation of a non-linear kernel and training
a support vector machine takes $ 3.9 $ hours, which is about six times longer than the time
needed to retrain the neural network with the feature map layer ($ 40 $ minutes).

A comparison of the results in Table \ref{tab:differentKernels}
reveals that the multichannel RBF-$ \chi^2 $ kernel as used in~\cite{wang2013dense}
does not achieve as good results as a simple $ \chi^2 $ kernel with feature concatenation.

\begin{table*}
\begin{center}
    \begin{tabularx}{\textwidth}{l X r r r r X r r r r}
        \toprule
                                         & & \multicolumn{4}{c}{concatenated}    & & \multicolumn{4}{c}{separate} \\
                                           \cmidrule(lr){3-6}                                    \cmidrule(lr){8-11}
                                         & &           &            & hist. & RBF-$ \chi^2 $ & &           &            & hist. & RBF-$ \chi^2 $ \\
        \multicolumn{2}{c}{feature map:}   & Hellinger & $ \chi^2 $ & int.  & Eq. \eqref{multichannelChiSquare} & & Hellinger & $ \chi^2 $ & int.  & Eq. \eqref{multichannelChiSquare} \\
        \midrule
        \textit{Olympic Sports}          \\
        kMeans + SVM                     & & $ 83.7 $  & $ 84.2 $   & $ 85.5 $ & $ 84.1 $  & & $ 85.9 $  & $ 87.1 $  & $ 86.6 $ & $ 84.4 $ \\
        neural network + SVM             & & $ 85.0 $  & $ 83.0 $   & $ 86.6 $ & $ 86.7 $  & & $ 86.4 $  & $ \mathbf{88.1} $  & $ 87.5 $ & $ 85.9 $ \\
        neural network (retrain top)     & & $ 86.6 $  & $ 85.8 $   & $ 85.8 $ & -         & & $ 78.5 $  & $ 83.5 $  & $ 84.1 $ & - \\
        \midrule
        \textit{J-HMDB}                  \\
        kMeans + SVM                     & & $ 57.1 $  & $ 56.0 $   & $ 56.8 $ & $ 56.6 $  & & $ 60.9 $  & $ 60.0 $  & $ 59.8 $ & $ 59.1 $ \\
        neural network + SVM             & & $ 59.4 $  & $ 58.5 $   & $ 58.3 $ & $ 57.6 $  & & $ \mathbf{62.8} $  & $ 62.2 $  & $ 61.9 $ & $ 61.9 $ \\
        neural network (retrain top)     & & $ 57.5 $  & $ 56.6 $   & $ 56.7 $ & -         & & $ 56.9 $  & $ 60.9 $  & $ 60.5 $ & - \\
        \midrule
        \textit{HMDB-51}                 \\
        kMeans + SVM                     & & $ 42.1 $  & $ 43.9 $   & $ 44.0 $ & $ 45.8 $  & & $ 51.4 $  & $ 52.4 $  & $ 51.5 $ & $ 52.2 $ \\
        neural network + SVM             & & $ 48.0 $  & $ 49.3 $   & $ 50.0 $ & $ 50.6 $  & & $ 53.6 $  & $ \mathbf{54.9} $  & $ 54.3 $ & $ 54.0 $ \\
        neural network (retrain top)     & & $ 44.0 $  & $ 47.3 $   & $ 47.4 $ & -         & & $ 47.9 $  & $ 50.2 $  & $ 50.0 $ & - \\
        \midrule
        \textit{UCF101}                  \\
        kMeans + SVM                     & & $ 72.2 $  & $ 73.1 $   & $ 72.2 $ & $ 67.8 $  & & $ 79.2 $  & $ 79.5 $  & $ 78.7 $ & $ 73.3 $ \\
        neural network + SVM             & & $ 78.8 $  & $ 78.6 $   & $ 77.7 $ & $ 73.3 $  & & $ 81.7 $  & $ \mathbf{81.9} $  & $ 81.1 $ & $ 76.9 $ \\
        neural network (retrain top)     & & $ 72.7 $  & $ 73.5 $   & $ 73.3 $ & -         & & $ 73.5 $  & $ 75.7 $  & $ 75.7 $ & - \\

        \bottomrule
    \end{tabularx}
\end{center}
\caption{Evaluation of different kernels for the kMeans baseline, the neural network approach with support vector machine classification,
         and the neural network including a feature map layer. For the latter, we use the \textit{retrain top} strategy.}
\label{tab:differentKernels}
\end{table*}

\subsection{Comparison to State of the Art}

\begin{table*}
\begin{center}
    \begin{tabularx}{\textwidth}{X r r r}
        \toprule
        Method                                                                     & HMDB-51    & & UCF101     \\
        \midrule
        \textit{Traditional models} & & \hspace{0.05\textwidth} & \\
        Improved DT + bag of words                                                 & $ 52.2\% $ & & $ 73.3\% $ \\
        Improved DT + Fisher vectors~\cite{wang2013action, wang2013thumos} (*)     & $ 57.2\% $ & & $ 85.9\% $ \\
        Improved DT + LLC                                                          & $ 50.8\% $ & & $ 71.9\% $ \\
        Stacked Fisher vectors~\cite{peng2014action} (*)                           & $ 66.8\% $ & & -          \\
        Multi-skip feature stacking~\cite{lan2014beyond} (*)                       & $ 65.4\% $ & & $ 89.1\% $ \\
        Super-sparse coding vector~\cite{yang2014action}                           & $ 53.9\% $ & & -          \\
        Motion-part regularization~\cite{ni2015motion} (*)                         & $ 65.5\% $ & & -          \\
        MoFap~\cite{wang2016mofap} (*)                                             & $ 61.7\% $ & & $ 88.3\% $ \\
        \midrule
        \textit{Neural networks} & & & \\
        Two-stream CNN~\cite{simonyan2014two} (**)                                 & $ 59.4\% $ & & $ 88.0\% $ \\
        Slow-fusion spatio-temporal CNN~\cite{karpathy2014large} (**)              & -          & & $ 65.4\% $ \\
        Composite LSTM~\cite{srivastava2015unsupervised}                           & $ 44.0\% $ & & $ 75.8\% $ \\
        TDD + improved DT with Fisher vectors~\cite{wang2015trajectorypooled} (*)  & $ 65.9\% $ & & $ 91.5\% $ \\
        factorized spatio-temporal CNN~\cite{sun2015factorized}                    & $ 59.1\% $ & & $ 88.1\% $ \\
        \midrule
        \textbf{Ours}                                                              & $ \mathbf{54.9\%} $ & & $ \mathbf{81.9\%} $ \\
        \bottomrule
    \end{tabularx}
\end{center}
\caption{Comparison of our model to published results on HMDB-51 and UCF101. We also provide results with bag-of-words and LLC encoding as a direct comparison to our method.
         Methods marked with (*) use Fisher vectors, those marked with (**) use additional training data.}
\label{tab:stateOfTheArt}
\end{table*}

In Table \ref{tab:stateOfTheArt}, our best results - a neural network without feature map layer
and a non-multichannel $ \chi^2 $ kernel - are compared to the state-of-the-art on HMDB-51 and UCF101.
Our approach outperforms other approaches based on bag-of-words, sparse coding~\cite{yang2014action},
locality constrained linear coding (LLC)~\cite{wang2010locality}, or neural networks~\cite{karpathy2014large,
srivastava2015unsupervised}. Particularly on UCF101, the improvement of $ 8\% $ compared to the original
bag-of-words based pipeline is remarkable. The approach~\cite{simonyan2014two} is not directly comparable
since the accuracy is mainly boosted by the use of additional training data. The approach of~\cite{sun2015factorized}
outperforms our method. This method uses a large convolutional neural network consisting of multiple
spatial convolution layers and a temporal convolution to enable modeling actions of different speeds.
Apart from that method, only the methods that use Fisher vectors achieve a better accuracy than our method.
However, extracting Fisher vectors is more expensive in terms of memory than a bag-of-words model. If Fisher
vectors are extracted per frame, the storage of the features would require around 1 TB for the $ 2.4 $ million
frames of UCF101 compared to 35 GB for our method. For applications with memory and runtime constraints,
our approach is a very useful alternative.



\section{Conclusion}

In this work, we have proposed a recurrent neural network that allows for
discriminative and supervised visual vocabulary generation. In contrast to
many existing coding and CNN based methods, our method can be applied on video
level directly.
Apart from kernel approximations via explicit feature maps, the network is
equivalent to the traditional bag-of-words approach and differs only in the
way it is trained.
Although the best results could be obtained using the neural network for
visual vocabulary generation and a support vector machine for classification,
we have shown that it is possible to also include the kernel and classification
steps into the network while retaining the performance of the original
bag-of-words model. Our model has been particularly beneficial for large scale
datasets. Moreover, it allows for a significant reduction in the amount of extracted
features, speeding up training time and inference without a considerable loss in
performance. Finally, our model can also be applied to other tasks like image
classification. The neural network proves to be competitive with other methods
that introduce additional constraints like sparsity.


\paragraph{Acknowledgments}
Authors acknowledge financial support by the ERC starting grant ARCA (677650).

\section*{References}

\bibliography{mybibfile}

\begin{thebibliography}{10}
\expandafter\ifx\csname url\endcsname\relax
  \def\url#1{\texttt{#1}}\fi
\expandafter\ifx\csname urlprefix\endcsname\relax\def\urlprefix{URL }\fi
\expandafter\ifx\csname href\endcsname\relax
  \def\href#1#2{#2} \def\path#1{#1}\fi

\bibitem{csurka2004visual}
G.~Csurka, C.~Dance, L.~Fan, J.~Willamowski, C.~Bray, Visual categorization
  with bags of keypoints, in: ECCV Workshop on statistical learning in computer
  vision, 2004.

\bibitem{sivic2005discovering}
J.~Sivic, B.~C. Russell, A.~A. Efros, A.~Zisserman, W.~T. Freeman, Discovering
  object categories in image collections, Tech. rep., Massachusetts Institute
  of Technology (2005).

\bibitem{zhang2007local}
J.~Zhang, M.~Marsza{\l}ek, S.~Lazebnik, C.~Schmid, Local features and kernels
  for classification of texture and object categories: A comprehensive study,
  International Journal on Computer Vision 73 (2007) 213--238.

\bibitem{VideoGoogle}
J.~Sivic, A.~Zisserman, Video google: a text retrieval approach to object
  matching in videos, in: Int. Conf. on Computer Vision, 2003, pp. 1470--1477.

\bibitem{krizhevsky2012imagenet}
A.~Krizhevsky, I.~Sutskever, G.~E. Hinton, Imagenet classification with deep
  convolutional neural networks, in: Advances in Neural Information Processing
  Systems, 2012, pp. 1097--1105.

\bibitem{perronnin2010improving}
F.~Perronnin, J.~S{\'a}nchez, T.~Mensink, Improving the {F}isher kernel for
  large-scale image classification, in: European Conf. on Computer Vision,
  2010, pp. 143--156.

\bibitem{wang2013dense}
H.~Wang, A.~Kl{\"a}ser, C.~Schmid, C.-L. Liu, Dense trajectories and motion
  boundary descriptors for action recognition, International Journal on
  Computer Vision 103 (2013) 60--79.

\bibitem{wang2013action}
H.~Wang, C.~Schmid, Action recognition with improved trajectories, in: Int.
  Conf. on Computer Vision, 2013, pp. 3551--3558.

\bibitem{taralova2014motion}
E.~H. Taralova, F.~De~la Torre, M.~Hebert, Motion words for videos, in:
  European Conf. on Computer Vision, 2014, pp. 725--740.

\bibitem{reddy2013recognizing}
K.~K. Reddy, M.~Shah, Recognizing 50 human action categories of web videos,
  Machine Vision and Applications 24 (2013) 971--981.

\bibitem{jhuang2013jhmdb}
H.~Jhuang, J.~Gall, S.~Zuffi, C.~Schmid, M.~J. Black, Towards understanding
  action recognition, in: Int. Conf. on Computer Vision, 2013, pp. 3192--3199.

\bibitem{peng2014bag}
X.~Peng, L.~Wang, X.~Wang, Y.~Qiao, Bag of visual words and fusion methods for
  action recognition: Comprehensive study and good practice, Computer Vision
  and Image Understanding.

\bibitem{richard2015bow}
A.~Richard, J.~Gall, A bow-equivalent recurrent neural network for action
  recognition, in: British Machine Vision Conference, 2015.

\bibitem{perronnin2006adapted}
F.~Perronnin, C.~Dance, G.~Csurka, M.~Bressan, Adapted vocabularies for generic
  visual categorization, in: European Conf. on Computer Vision, 2006, pp.
  464--475.

\bibitem{cai2010learning}
H.~Cai, F.~Yan, K.~Mikolajczyk, Learning weights for codebook in image
  classification and retrieval, in: IEEE Conf. on Computer Vision and Pattern
  Recognition, 2010, pp. 2320--2327.

\bibitem{lian2010probabilistic}
X.-C. Lian, Z.~Li, C.~Wang, B.-L. Lu, L.~Zhang, Probabilistic models for
  supervised dictionary learning, in: IEEE Conf. on Computer Vision and Pattern
  Recognition, 2010, pp. 2305--2312.

\bibitem{yang2009linear}
J.~Yang, K.~Yu, Y.~Gong, T.~Huang, Linear spatial pyramid matching using sparse
  coding for image classification, in: IEEE Conf. on Computer Vision and
  Pattern Recognition, 2009, pp. 1794--1801.

\bibitem{wang2010locality}
J.~Wang, J.~Yang, K.~Yu, F.~Lv, T.~Huang, Y.~Gong, Locality-constrained linear
  coding for image classification, in: IEEE Conf. on Computer Vision and
  Pattern Recognition, 2010, pp. 3360--3367.

\bibitem{goh2012unsupervised}
H.~Goh, N.~Thome, M.~Cord, J.-H. Lim, Unsupervised and supervised visual codes
  with restricted boltzmann machines, in: European Conf. on Computer Vision,
  2012, pp. 298--311.

\bibitem{boureau2010learning}
Y.-L. Boureau, F.~Bach, Y.~LeCun, J.~Ponce, Learning mid-level features for
  recognition, in: IEEE Conf. on Computer Vision and Pattern Recognition, 2010,
  pp. 2559--2566.

\bibitem{jegou2012aggregating}
H.~J{\'e}gou, F.~Perronnin, M.~Douze, J.~Sanchez, P.~Perez, C.~Schmid,
  Aggregating local image descriptors into compact codes, IEEE Transactions on
  Pattern Analysis and Machine Intelligence 34 (2012) 1704--1716.

\bibitem{peng2014action}
X.~Peng, C.~Zou, Y.~Qiao, Q.~Peng, Action recognition with stacked {F}isher
  vectors, in: European Conf. on Computer Vision, 2014, pp. 581--595.

\bibitem{oneata2013action}
D.~Oneata, J.~Verbeek, C.~Schmid, Action and event recognition with {F}isher
  vectors on a compact feature set, in: Int. Conf. on Computer Vision, 2013,
  pp. 1817--1824.

\bibitem{simonyan2013deep}
K.~Simonyan, A.~Vedaldi, A.~Zisserman, Deep {F}isher networks for large-scale
  image classification, in: Advances in Neural Information Processing Systems,
  2013, pp. 163--171.

\bibitem{wang2016mofap}
L.~Wang, Y.~Qiao, X.~Tang, Mofap: A multi-level representation for action
  recognition, International Journal on Computer Vision 119~(3) (2016)
  254--271.

\bibitem{ni2015motion}
B.~Ni, P.~Moulin, X.~Yang, S.~Yan, Motion part regularization: Improving action
  recognition via trajectory selection, in: IEEE Conf. on Computer Vision and
  Pattern Recognition, 2015, pp. 3698--3706.

\bibitem{peng2014boosting}
X.~Peng, L.~Wang, Y.~Qiao, Q.~Peng, Boosting vlad with supervised dictionary
  learning and high-order statistics, in: European Conf. on Computer Vision,
  2014, pp. 660--674.

\bibitem{sydorov2014deep}
V.~Sydorov, M.~Sakurada, C.~H. Lampert, Deep {F}isher kernels--end to end
  learning of the {F}isher kernel {G}mm parameters, in: IEEE Conf. on Computer
  Vision and Pattern Recognition, 2014, pp. 1402--1409.

\bibitem{karpathy2014large}
A.~Karpathy, G.~Toderici, S.~Shetty, T.~Leung, R.~Sukthankar, L.~Fei-Fei,
  Large-scale video classification with convolutional neural networks, in: IEEE
  Conf. on Computer Vision and Pattern Recognition, 2014, pp. 1725--1732.

\bibitem{simonyan2014two}
K.~Simonyan, A.~Zisserman, Two-stream convolutional networks for action
  recognition in videos, in: Advances in Neural Information Processing Systems,
  2014, pp. 568--576.

\bibitem{donahue2015long}
J.~Donahue, L.~A. Hendricks, S.~Guadarrama, M.~Rohrbach, S.~Venugopalan,
  K.~Saenko, T.~Darrell, Long-term recurrent convolutional networks for visual
  recognition and description, in: IEEE Conf. on Computer Vision and Pattern
  Recognition, 2015, pp. 2625--2634.

\bibitem{srivastava2015unsupervised}
N.~Srivastava, E.~Mansimov, R.~Salakhutdinov, Unsupervised learning of video
  representations using {LSTM}s, in: Int. Conf. on Machine Learning, 2015.

\bibitem{jain2015objects}
M.~Jain, J.~C. van Gemert, C.~G.~M. Snoek, What do 15,000 object categories
  tell us about classifying and localizing actions?, in: IEEE Conf. on Computer
  Vision and Pattern Recognition, 2015, pp. 46--55.

\bibitem{wang2015tdd}
L.~Wang, Y.~Qiao, X.~Tang, Action recognition with trajectory-pooled
  deep-convolutional descriptors, in: IEEE Conf. on Computer Vision and Pattern
  Recognition, 2015, pp. 4305--4314.

\bibitem{macherey2003comparative}
W.~Macherey, H.~Ney, A comparative study on maximum entropy and discriminative
  training for acoustic modeling in automatic speech recognition., in:
  Interspeech, 2003.

\bibitem{heigold2007equivalence}
G.~Heigold, R.~Schl{\"u}ter, H.~Ney, On the equivalence of {G}aussian {HMM} and
  {G}aussian {HMM}-like hidden conditional random fields., in: Interspeech,
  2007, pp. 1721--1724.

\bibitem{vedaldi2012efficient}
A.~Vedaldi, A.~Zisserman, Efficient additive kernels via explicit feature maps,
  IEEE Transactions on Pattern Analysis and Machine Intelligence (2012)
  480--492.

\bibitem{riedmiller1993direct}
M.~Riedmiller, H.~Braun, A direct adaptive method for faster backpropagation
  learning: The rprop algorithm, in: IEEE Int. Conf. on Neural Networks, 1993,
  pp. 586--591.

\bibitem{werbos1990backpropagation}
P.~J. Werbos, Backpropagation through time: what it does and how to do it,
  Proceedings of the IEEE 78 (1990) 1550--1560.

\bibitem{rumelhart1988learning}
D.~E. Rumelhart, G.~E. Hinton, R.~J. Williams, Learning representations by
  back-propagating errors, Nature 323 (1986) 533--536.

\bibitem{chang2011libsvm}
C.-C. Chang, C.-J. Lin, Libsvm: a library for support vector machines, ACM
  Transactions on Intelligent Systems and Technology 2 (2011) 1--27.

\bibitem{niebles2010modeling}
J.~C. Niebles, C.-W. Chen, L.~Fei-Fei, Modeling temporal structure of
  decomposable motion segments for activity classification, in: European Conf.
  on Computer Vision, 2010, pp. 392--405.

\bibitem{kuehne11hmdb}
H.~Kuehne, H.~Jhuang, E.~Garrote, T.~Poggio, T.~Serre, {HMDB}: a large video
  database for human motion recognition, in: Int. Conf. on Computer Vision,
  2011, pp. 2556--2563.

\bibitem{soomro2012ucf101}
K.~Soomro, A.~R. Zamir, M.~Shah, Ucf101: A dataset of 101 human actions classes
  from videos in the wild, arXiv preprint arXiv:1212.0402.

\bibitem{he2015deep}
K.~He, X.~Zhang, S.~Ren, J.~Sun, Deep residual learning for image recognition,
  arXiv preprint arXiv:1512.03385.

\bibitem{cho2014learning}
K.~Cho, B.~Van~Merri{\"{e}}nboer, {\c C}.~G{\"{u}}l{\c c}ehre, D.~Bahdanau,
  F.~Bougares, H.~Schwenk, Y.~Bengio, Learning phrase representations using rnn
  encoder--decoder for statistical machine translation, in: Conf. on Empirical
  Methods in Natural Language Processing, 2014, pp. 1724--1734.

\bibitem{bahdanau2015neural}
D.~Bahdanau, K.~Cho, Y.~Bengio, Neural machine translation by jointly learning
  to align and translate, in: Int. Conf. on Learning Representations, 2015.

\bibitem{lazebnik2006beyond}
S.~Lazebnik, C.~Schmid, J.~Ponce, Beyond bags of features: Spatial pyramid
  matching for recognizing natural scene categories, in: IEEE Conf. on Computer
  Vision and Pattern Recognition, 2006, pp. 2169--2178.

\bibitem{van2010visual}
J.~van Gemert, C.~Veenman, A.~Smeulders, J.-M. Geusebroek, Visual word
  ambiguity, IEEE Transactions on Pattern Analysis and Machine Intelligence 32
  (2010) 1271--1283.

\bibitem{liu2011defense}
L.~Liu, L.~Wang, X.~Liu, In defense of soft-assignment coding, in: Int. Conf.
  on Computer Vision, 2011, pp. 2486--2493.

\bibitem{boureau2011ask}
Y.-L. Boureau, N.~Le~Roux, F.~Bach, J.~Ponce, Y.~LeCun, Ask the locals:
  multi-way local pooling for image recognition, in: Int. Conf. on Computer
  Vision, 2011, pp. 2651--2658.

\bibitem{fei2007learning}
L.~Fei-Fei, R.~Fergus, P.~Perona, Learning generative visual models from few
  training examples: An incremental bayesian approach tested on 101 object
  categories, Computer Vision and Image Understanding (2007) 59--70.

\bibitem{fei2005bayesian}
L.~Fei-Fei, P.~Perona, A bayesian hierarchical model for learning natural scene
  categories, in: IEEE Conf. on Computer Vision and Pattern Recognition, 2005,
  pp. 524--531.

\bibitem{wang2013thumos}
H.~Wang, C.~Schmid, {LEAR-INRIA} submission for the thumos workshop, in: ICCV
  Workshop on Action Recognition with a Large Number of Classes, 2013.

\bibitem{lan2014beyond}
Z.~Lan, M.~Lin, X.~Li, A.~G. Hauptmann, B.~Raj, Beyond {G}aussian pyramid:
  Multi-skip feature stacking for action recognition, in: IEEE Conf. on
  Computer Vision and Pattern Recognition, 2015, pp. 204--212.

\bibitem{yang2014action}
X.~Yang, Y.~Tian, Action recognition using super sparse coding vector with
  spatio-temporal awareness, in: European Conf. on Computer Vision, 2014.

\bibitem{wang2015trajectorypooled}
L.~Wang, Y.~Qiao, X.~Tang, Action recognition with trajectory-pooled
  deep-convolutional descriptors, in: IEEE Conf. on Computer Vision and Pattern
  Recognition, 2015, pp. 4305--4314.

\bibitem{sun2015factorized}
L.~Sun, K.~Jia, D.-Y. Yeung, B.~Shi, Human action recognition using factorized
  spatio-temporal convolutional networks, in: Int. Conf. on Computer Vision,
  2015.

\end{thebibliography}

\end{document}